\documentclass{article}
\usepackage{iftex}
\ifPDFTeX
  \usepackage[utf8]{inputenc}
\fi
\usepackage[T1]{fontenc}
\usepackage{genai4health_2026}
\usepackage{hyperref}
\usepackage{url}
\usepackage{booktabs}
\usepackage{amsmath,amssymb}
\usepackage{graphicx}
\hypersetup{hidelinks}
\usepackage{xcolor}
\usepackage{placeins}
\usepackage[utf8]{inputenc}
\usepackage[T1]{fontenc}
\usepackage{hyperref}
\usepackage{url}
\usepackage{booktabs}
\usepackage{multirow}
\usepackage{makecell}
\usepackage{amsfonts}
\usepackage{amsmath}
\usepackage{amssymb}
\usepackage{nicefrac}
\usepackage{microtype}
\usepackage{xcolor}
\usepackage{inconsolata}
\usepackage{graphicx}

\title{\textcolor{purple}{MSM-Mem}: A Universal \textcolor{purple}{M}edical \textcolor{purple}{S}tructured \textcolor{purple}{M}ultimodal \textcolor{purple}{Mem}ory Framework for Medical AI Agents}

\author{%
  Md Asaduzzaman Jabin \\
  College of Engineering, University of Georgia \\
  Athens, GA, USA \\
  \texttt{mj71006@uga.edu}
  \And
  Khoa Le \\
  College of Engineering, University of Georgia \\
  Athens, GA, USA \\
  \texttt{Khoa.Le1@uga.edu}
  \And
  Lin Zhao\\
  Department of Biomedical Engineering, New Jersey Institute of Technology \\
  Newark, NJ, USA \\
  \texttt{lin.zhao.1@njit.edu}
  \And
  Tianming Liu\thanks{Corresponding author.} \\
  College of Engineering, University of Georgia \\
  Athens, GA, USA \\
  \texttt{tliu@uga.edu}
}

\begin{document}
\maketitle

\begin{abstract}
Clinical decision-making is inherently experience-driven: physicians progressively refine their reasoning by synthesizing patient history, multimodal observations, and prior diagnostic experiences across interactions. In contrast, current multimodal large language model (MLLM)-based medical AI agents largely operate as stateless inference systems, generating decisions independently for each interaction without retaining or internalizing experiential knowledge. This discrepancy limits their ability to progressively improve reasoning reliability through usage and adapt to longitudinal patient contexts in real-world clinical workflows. In this study, we propose Medical Structured Multimodal Memory (MSM-Mem), an agentic memory framework that enables medical AI agents to evolve through accumulated clinical experiences. MSM-Mem organizes heterogeneous clinical experiences into semantic, episodic, and visual memory and incrementally updates them during inference, allowing the agent to retrieve prior experiences to inform current reasoning and progressively refine decision-making over time. Evaluations on MoE-LLaVA backbones demonstrate consistent performance improvements with further gains observed through continued usage. In general, MSM-Mem offers a viable pathway toward medical AI agents capable of evolving their reasoning competence in a manner analogous to the way clinicians learn from practice over time.
\end{abstract}

\begin{figure}[t]
  \centering
  \includegraphics[width=0.9\linewidth]{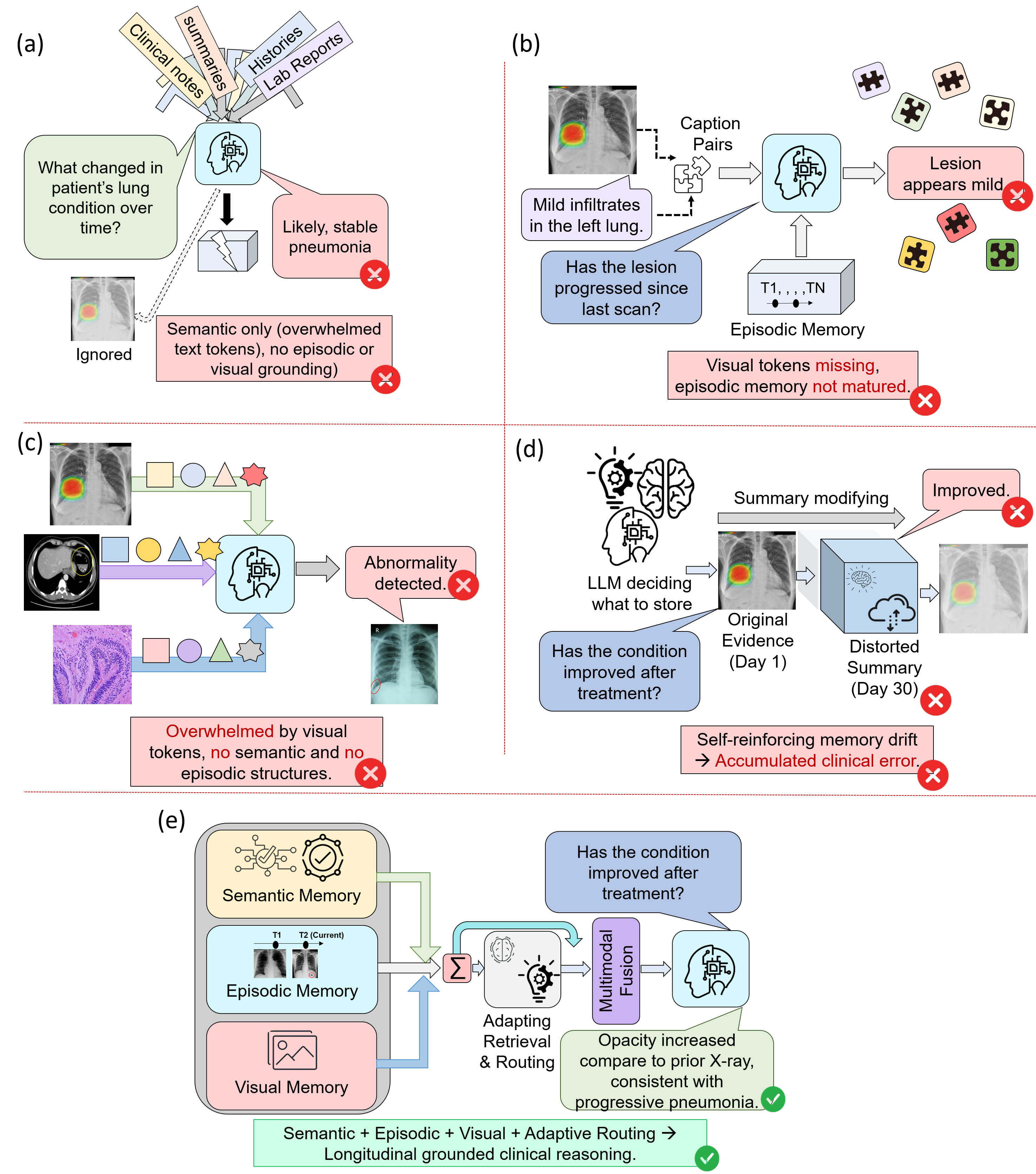}
  \caption{Conceptual comparison of agent memory- (a) Overwhelmed by massive text tokens \cite{terranova_evaluating_2025}, \cite{bonnici_multimodal_2016}; (b) Image-caption pair with immature semantic and episodic structures \cite{selivanov_medical_2023}, \cite{beddiar_automatic_2023}; (c) Visual-only methods lacks semantic and episodic structures \cite{zhang_integrating_2022}, \cite{davis_visual_2020}; (d) LLM-curated approach with uncontrolled and hallucinated memory \cite{lin_llm-based_2025}, \cite{savage_large_2025}; and (e) MSM-Mem (Ours) with 3 persistent memories with efficient retrieval and adaptive routing.}
  \label{fig:fig1}
\end{figure}

\begin{figure}[t]
  \centering
  \includegraphics[width=0.92\linewidth]{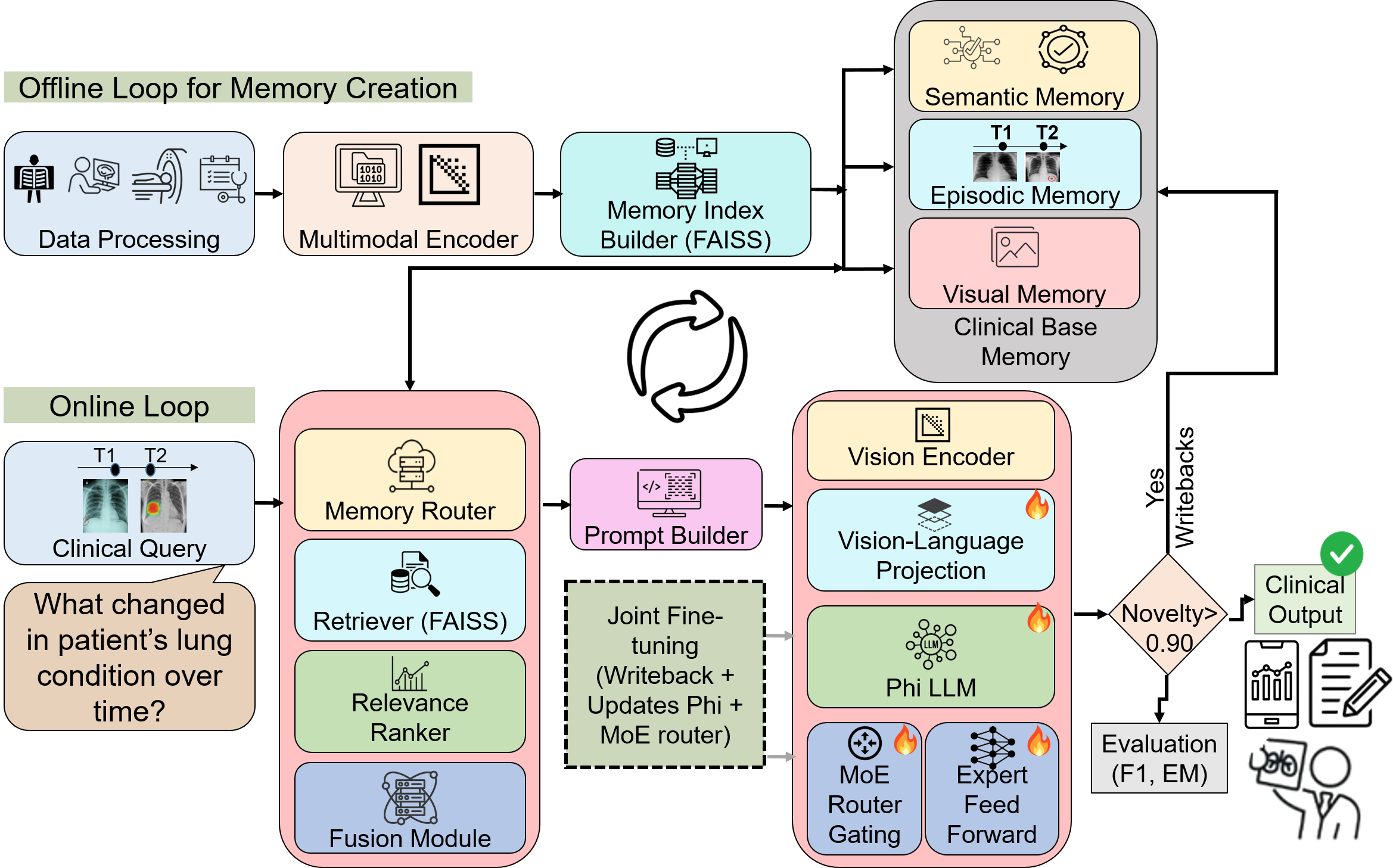} \hfill
  \caption {MSM-Mem agent self-evolving architecture- (a) Offline loop- encoding, memory builder, build world memory; and (b) Online loop- run queries, retrieve base and fused writeback stages, accessing fine-tuned foundation model, and adaptive routing.}
  \label{fig:fig2}
\end{figure}

\begin{figure}[t]
  \centering
  \includegraphics[width=0.82\linewidth]{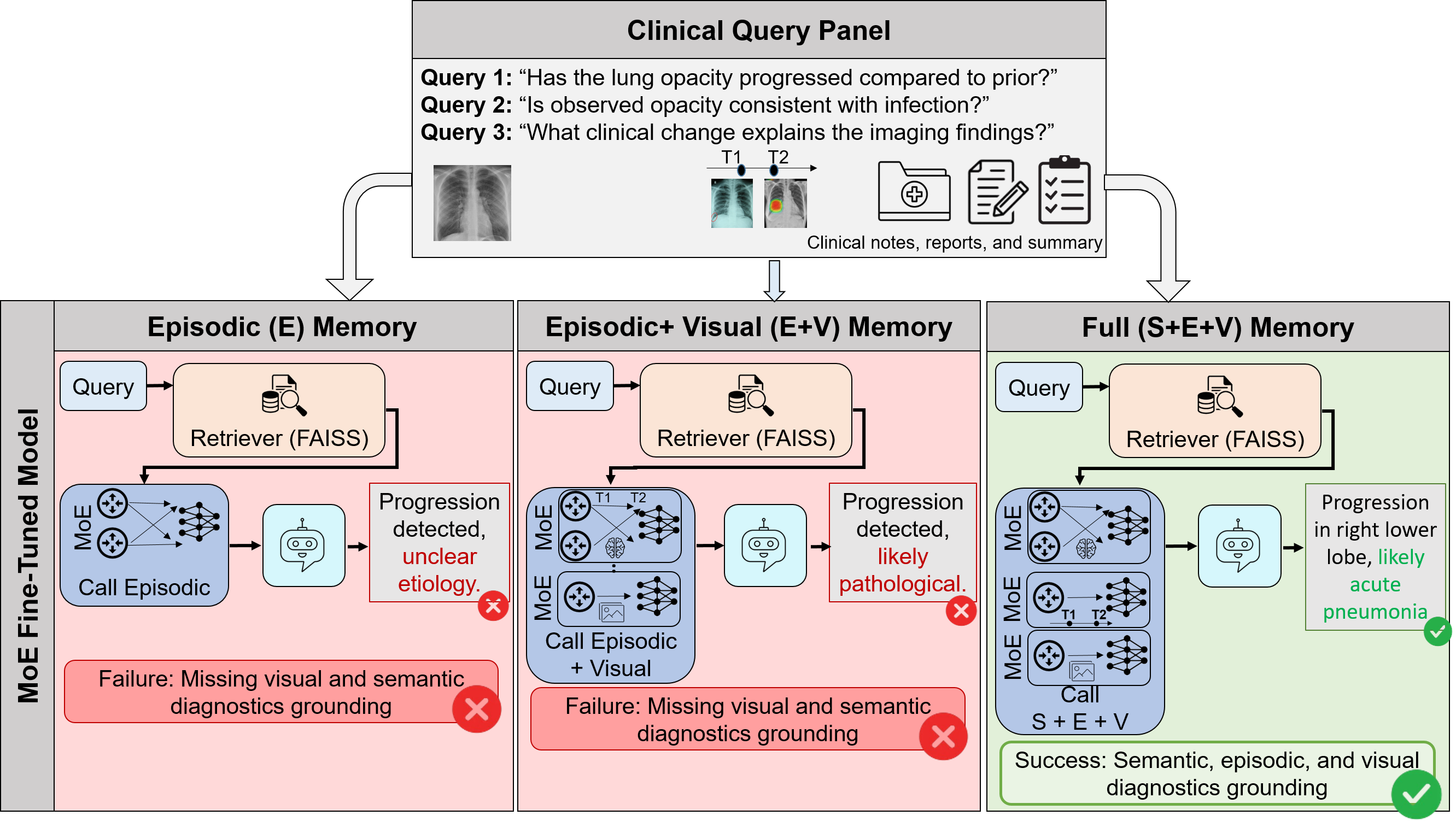} \hfill
  \caption {The impact of writeback phases at best performing memory modality (E, E+V, S+E+V) for MoE fine-tuned (MoE-FT) model. (Here, E = episodic, S = semantic, and V = visual memory).}
  \label{fig:fig3}
\end{figure}

\begin{figure}[t]
  \centering
  \includegraphics[width=0.40\linewidth]{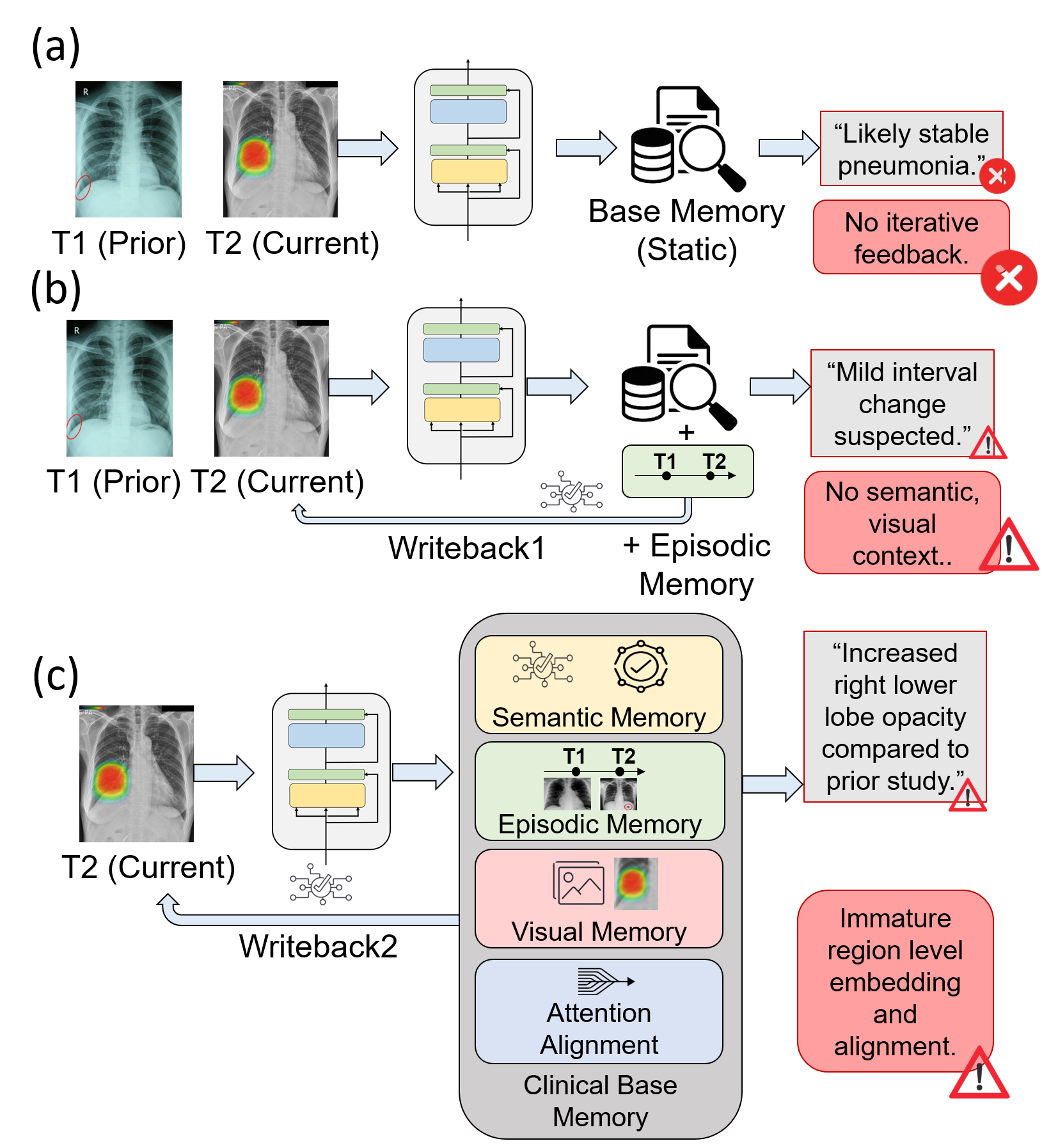} \hfill
  \includegraphics[width=0.40\linewidth]{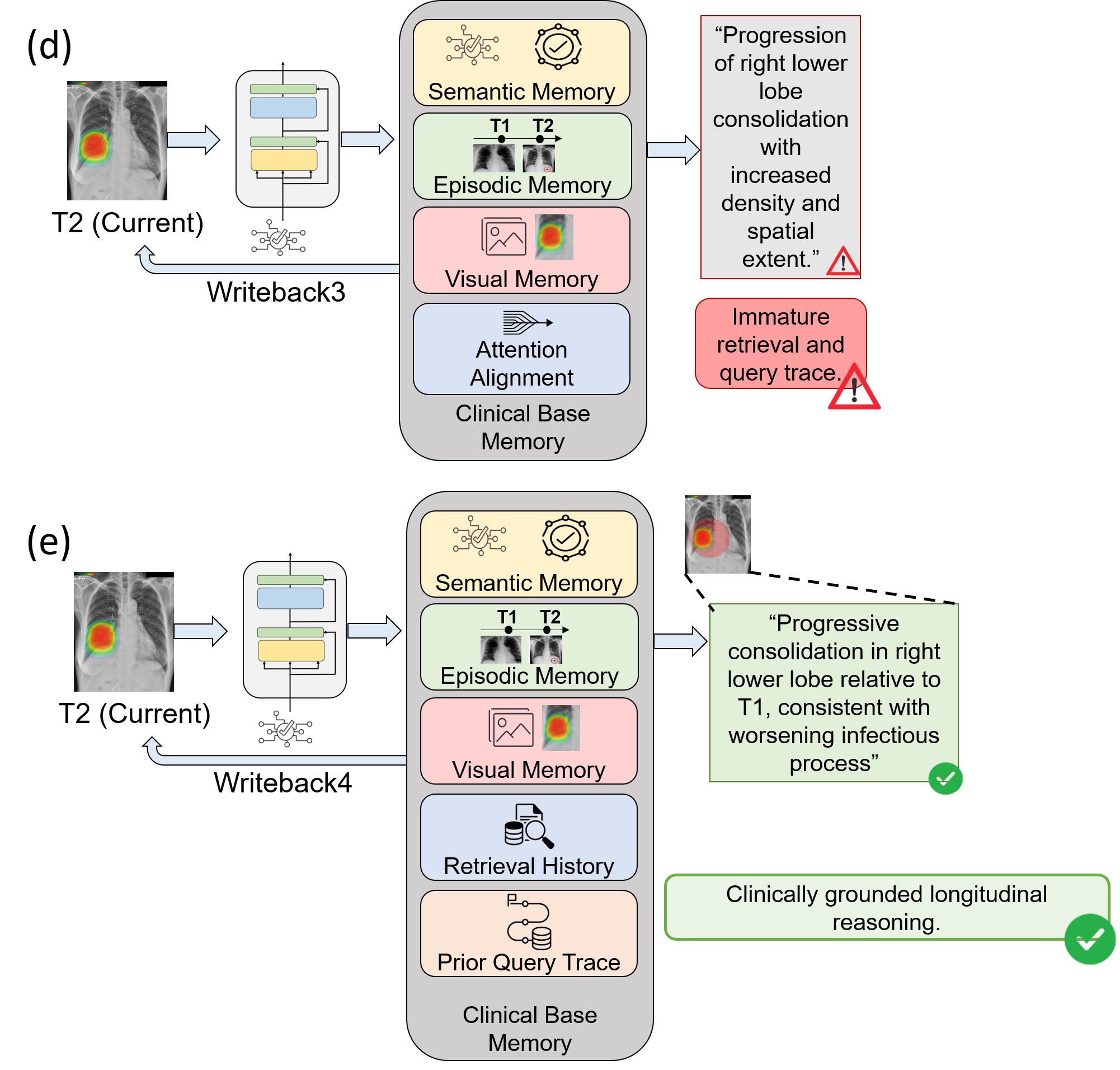}
  \caption {The impact of continuous memory improvement with writebacks (from base memory to WB4) for MoE fine-tuned (MoE-FT) model.}
  \label{fig:fig4}
\end{figure}

\begin{figure}[t]
  \centering
  \includegraphics[width=0.38\linewidth]{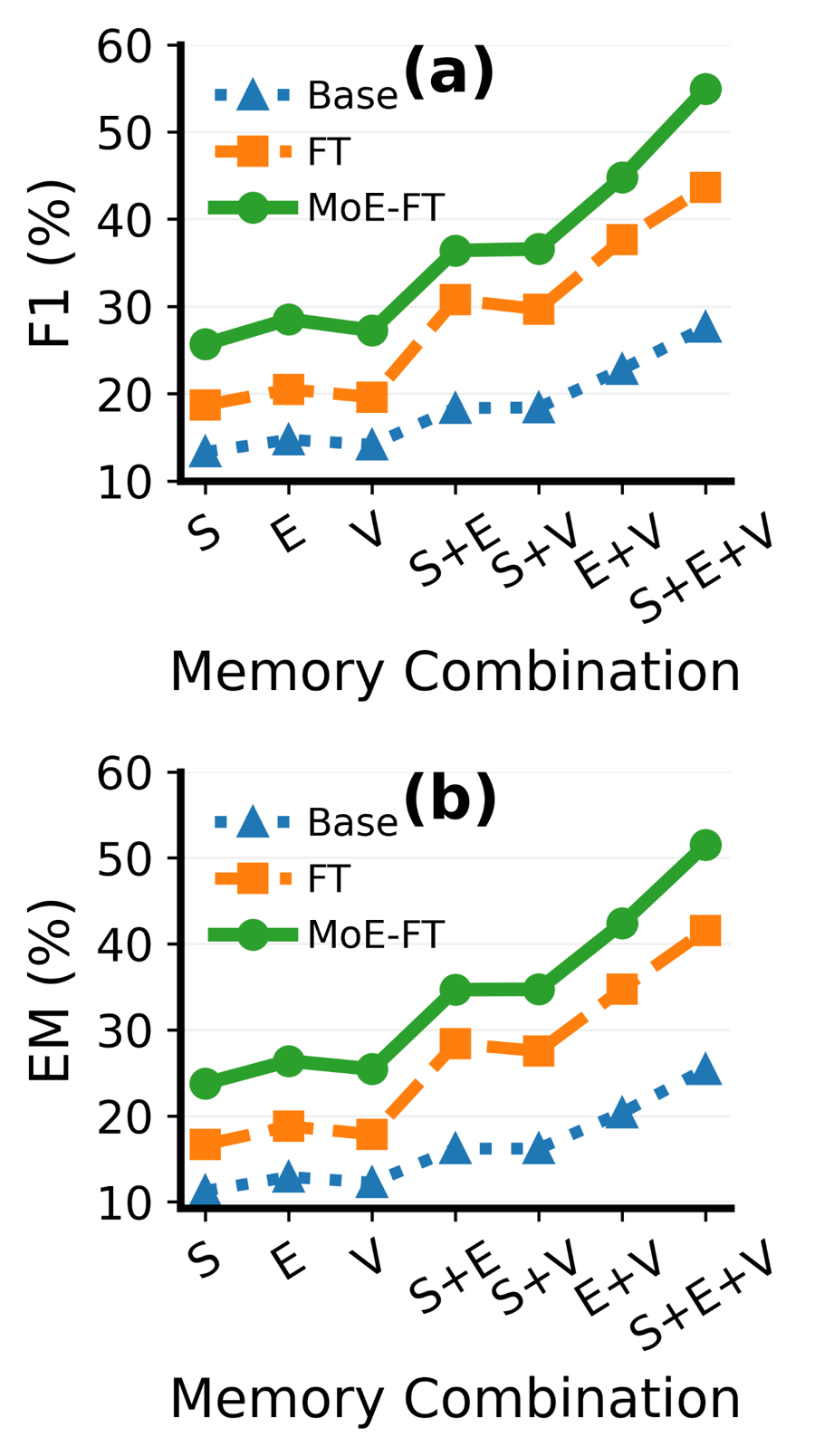}
  \caption{The memory combination’s impact at final memory phase (WB4) with metrics- (a) F1 impression average (F1); and (b) Exact match average (EM).}
  \label{fig:fig5}
\end{figure}

\begin{figure}[t]
  \centering
  \includegraphics[width=0.36\linewidth]{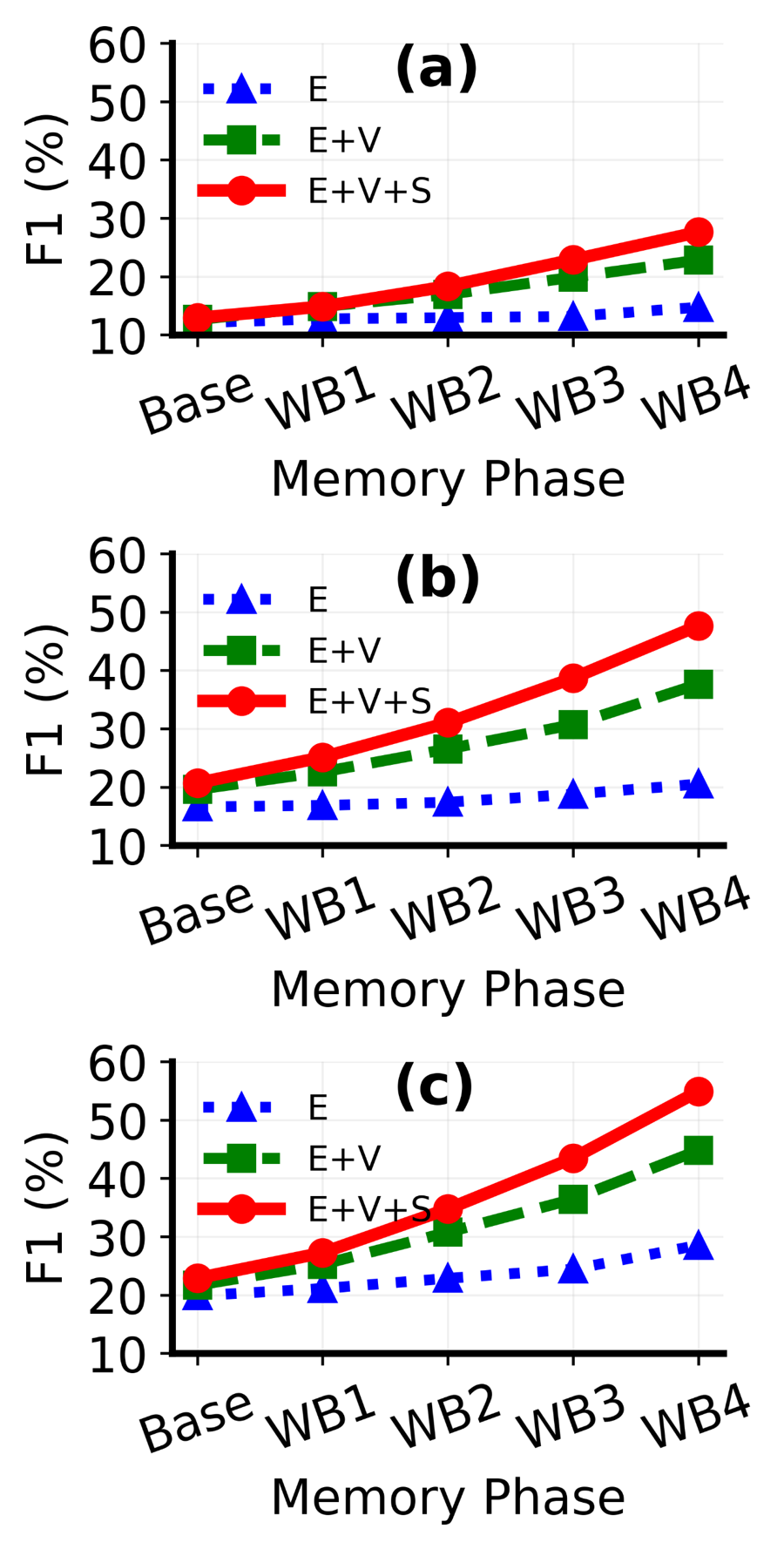}
  \caption{The impact of writeback phases at the best-performing memory modality (E, E+V, S+E+V) for each model checkpoint: (a) for the base model; (b) for the regular fine-tuned (FT) model; and (c) for the MoE fine-tuned (MoE-FT) model.}
  \label{fig:fig6}
\end{figure}

\begin{figure}[t]
  \centering
  \includegraphics[width=0.36\linewidth]{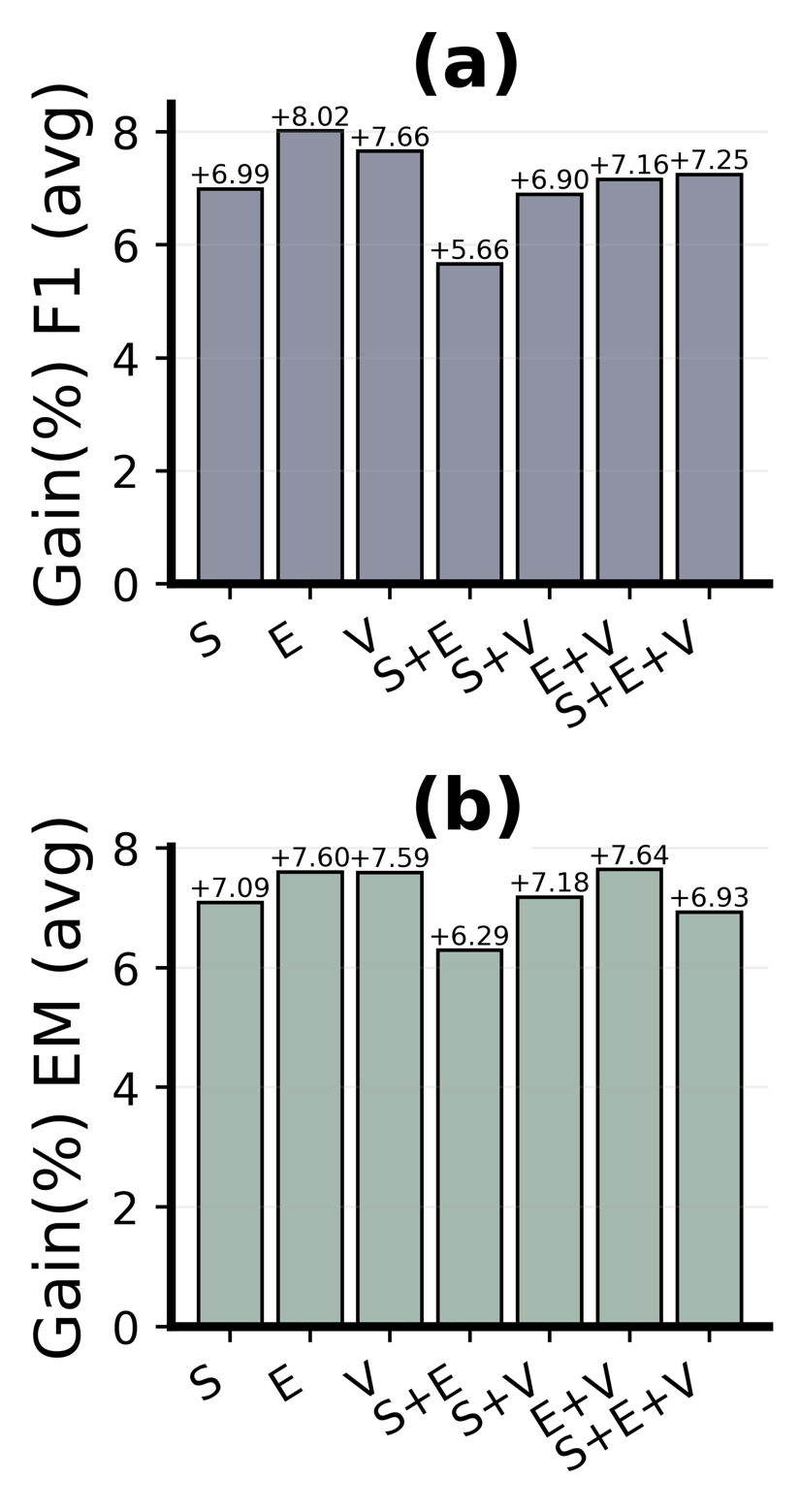}
  \caption{The comparison of MoE-FT vs. Non-MoE-FT (regular FT) at all memory modalities at best memory phase (WB4).}
  \label{fig:fig7}
\end{figure}

\section{Introduction}

Multimodal large language models (MLLMs) have recently demonstrated significant potential for medical and clinical applications \citet{maity_large_2025} by enabling joint reasoning over heterogeneous data sources such as radiological images, clinical notes, laboratory results, and structured patient records. By synthesizing multimodal evidence with medical knowledge representations, MLLMs can serve as the foundation for intelligent medical agents capable of assisting in various tasks such as report generation, clinical triage, disease monitoring, and decision support \cite{zhang_integrating_2022}. However, unlike clinicians whose reasoning is grounded in medical knowledge, patient-specific multimodal evidence, and accumulated experiential insights over time, current medical agents often generate hallucinated or clinically irrelevant information due to the absence of knowledge-, evidence-, and experience-informed contextual grounding, thereby limiting their effectiveness and reliability in real-world clinical workflows \cite{davis_visual_2020}.

Retrieval-augmented generation (RAG)~\cite{lewis_retrieval-augmented_2021} enhances context-grounded reasoning by retrieving relevant clinical documents or domain references at inference time. However, conventional RAG primarily performs external knowledge lookup and offers limited support for accumulating and internalizing experiential knowledge derived from prior clinical decisions or interactions, preventing iterative refinement through longitudinal use~\cite{ocheSystematicReviewKey2025,liu_challenges_2025}. Agentic memory mechanisms have been proposed to mitigate this statelessness by persistently storing and updating information across encounters~\cite{muksimova_multi-modal_2025}; however, existing approaches remain constrained in preserving modality-specific evidence over extended clinical contexts~\cite{terranova_evaluating_2025,lin_llm-based_2025}. 

In particular, semantic-only memories compress multimodal cues only into text summaries, often discarding fine-grained visual and episodic details and weakening evidence grounding~\cite{bonnici_multimodal_2016}; image–caption memories rely on descriptive proxies that can amplify caption bias and erode temporal consistency~\cite{reale-nosei_vision_2024,selivanov_medical_2023}; vision-only memories retain compact embeddings yet lack the semantic and temporal scaffolding necessary for clinical interpretation~\cite{bahleArchitectureInteractionVisual2018}; and LLM-curated memory construction may introduce uncontrolled abstraction and memory drift~\cite{pulipaka_persistbench_2026}. Collectively, these limitations indicate that current RAG and memory paradigms still fall short of maintaining modality-specific, experience-informed evidence required for grounded and consistent reasoning in complex longitudinal workflows~\cite{lin_llm-based_2025}. All those problems are depicted in Figure \ref{fig:fig1}.

To address these limitations, we propose a Medical Structured Multimodal Memory (MSM-Mem) framework, which decomposes multimodal clinical evidence into complementary semantic, episodic, and visual memory components, which respectively represent structured clinical knowledge, temporally indexed patient-specific observations, and fine-grained imaging evidence. New clinical evidence and decision-relevant outcomes are incrementally written back to the corresponding memory modules. This process accumulates experience across longitudinal interactions, thereby supporting progressive self-improvement over time. We evaluate our MSM-Mem framework on a MoE-based vision–language model backbone, MoE-LLaVA \cite{lin_moe-llava_2024}. Experimental results demonstrate that MSM-Mem consistently improves downstream task performance and reduces hallucinated outputs for both off-the-shelf and medically finetuned backbones. Incremental memory updates enable the agent to accumulate experience over time and progressively refine its reasoning capability.

\textbf{Medical AI Agent: }Contemporary medical agents \cite{fallahpourMedRAXMedicalReasoning2025}, \cite{yangLungNoduleAgentCollaborativeMultiAgent2025} have progressed to reason over complex multimodal cues by employing expanded token window. However, those models are tuned on generic data, persistently shows substantial computation cost, no adaptivity, shows semantic and episodic error, and not tuned especially to handle critical intelligence systems in clinical applications. Moreover, those model generally process every cue, therefore become inferior for queries targeted for deep clinical reasoning and disease diagnosis \cite{cao_development_2025}. On the other hand, memoryless AI agents are constrained by limited-term reasoning, knowledge inconsistency, constrained adaptability, and discontinuity of context switching \cite{wedelContextualMemoryIntelligence2025}. Without having permanent memory, a medical agent will fail to hold the preceding subject's record, the temporal disease progression, and diagnostic evidences which causes ununiform interpretation and lessened clinical reliability \cite{cuiTIMERTemporalInstruction2025}.

\textbf{Agentic Memory: }Recently, techniques such as, retrieval augmented generation (RAG) has used in building biomedical agents \cite{wuMedicalGraphRAG2025} that retrieves relevant information from external sources and fitting into a small token window (<3000 tokens/query) resulting into lacking of prior complex experimental knowledge extraction and temporal grounding. Later, several approaches are introduced by forming knowledge graph \cite{rezaeiAgenticMedicalKnowledge2025} to transcribe multimodal clinical cues that fails in less sparse coverage, resource allocation, temporal progression, and longitudinal reasoning. To enhance, persistent memory \cite{joUnderstandingImpactLongTerm2024}, \cite{wangMemaLearningMemory2025}, \cite{xuAMEMAgenticMemory2025} is introduced to deal with temporal grounding and episodic reasoning. However, current studies still grapple to entirely incorporate multimodal temporal cues, dynamic retrieval, choosing most fit foundation models and adaptive routing of LLM experts.

% \begin{figure}[t]
%   \includegraphics[width=1]{Fig-2.png} \hfill
%   \caption{A figure with a caption that runs for more than one line.
%     Example image is usually available through the \texttt{mwe} package
%     without even mentioning it in the preamble.}
%   \label{fig:fig1}
% \end{figure}

% \begin{figure}
% \includegraphics[width=\textwidth]{Fig-2.png}
% \caption{MSM-Mem agent self-evolving architecture- (a) Offline loop- encoding, memory builder, build world memory; and (b) Online loop- run queries, retrieve base and fused writeback stages, accessing fine-tuned foundation model, and adaptive routing.}
% \label{fig:fig1}

\section{Methods}

\subsection{Overview}
The overall MSM-Mem framework of MSM-Mem is illustrated in Figure~\ref{fig:fig2}. We propose MSM-Mem multimodal architecture designed for multimodal semantic insights, temporal cues, and long-term reasoning in biomedical applications. The pipeline consists of 5 stages, including offline, online, efficient retrieval, adaptive routing loop, and query inference (refer to Figure ~\ref{fig:fig2}). The offline loop is divided into multimodal encoders, a memory index builder, and multimodal base memory construction, which comprises semantic, episodic, and visual features, followed by sequential writebacks. The base memory in the offline loop is built over a large heterogeneous dataset and is built only once during system setup. The online loop comprises a unified retriever, prompt builder, and reasoning model with adaptive routing. The online loop is gradually enhancing the world's memory from time to time if new heterogeneous data occurs. This ensures that the inference becomes almost identical to the ground truth incrementally through continuous write-back phases. However, the retrieval consists of relevant memory modality, searches previous episodic embedding, re-ranking, and an efficient fusion module. And, the adaptive routing chooses a mixture of experts (MoE) efficiently based on ranking and cosine-similarity of memory features. Finally, each query is fed into the multimodal foundation agent by utilizing retrieved memory, cue history, and prompts to generate a final response or report. An intensive ablation study is done to show the performance of the proposed agent under various combinations of models and memory embedding. Appendix \ref{sec:appendix} shows the mathematical explanation of the memory in more detail.
% -------------------------------------
\subsection{Memory Construction}
\label{sec:memory}
Clinical decision is based on patient history, the current observations, and the relevant clinical experiences. MSM-Mem factorizes these complementary evidence types into structured memory components and maintains a structured memory store defined as $\mathcal{M} = \{M_S, M_E, M_V\}$, where $M_S$ is semantic memory, $M_E$ is episodic memory, and $M_V$ is visual memory. 

\textbf{Episodic Memory.}
A central objective of this work is to model the process by which clinicians progressively refine diagnostic decisions through accumulated experience across repeated patient encounters. To simulate this interaction-driven learning mechanism, MSM-Mem represents prior patient–clinician interactions as episodic memory, capturing how observed evidence was previously interpreted and translated into downstream decisions. Each interaction is stored as an episodic record:
$
e_i =
(\text{subj}_i,\text{epi}_i, t_i, z_i, q_i, a_i,d_i, \text{vis\_path}_i, \text{vis\_desc}_i)$,
% \begin{equation}
% e_i =
% (\text{subj}_i,\text{epi}_i, t_i, z_i, q_i, a_i,d_i, \text{vis\_path}_i, \text{vis\_desc}_i)
% \end{equation}
\noindent where $\text{subj}_i$ and $\text{epi}_i$ denote the subject and interaction identifiers, respectively, $t_i$ is the interaction timestamp, $z_i \in \mathbb{R}^d$ is the visual embedding associated with the observed input for similarity-based query matching, $q_i$ is the clinical query, $a_i$ is the generated response or impression, $d_i$ is the corresponding decision outcome, and $\text{vis\_path}_i$ and $\text{vis\_desc}_i$ denote the associated observation reference (e.g., image path) and its textual description. Note that to ensure consistent similarity-based retrieval, stored visual embeddings are normalized prior to memory write-back as $\hat{z}_i = z_i/\|z_i\|_2$.

\textbf{Visual Memory.}
To preserve the perceptual evidence independently of interaction-level decision context, MSM-Mem maintains a visual memory module that stores compact representations of previously encountered visual observations for prototype-level retrieval and comparison: \textit{$v_i$ =($\text{subj}_i$, $\text{epi}_i$, $\text{vis}_i$, $t_i$, $z_i$,$\text{vis\_tag}_i$, $\text{vis\_path}_i$)},
% \begin{equation}
% v_i =(\text{subj}_i, \text{epi}_i, \text{vis}_i, t_i, z_i,\text{vis\_tag}_i, \text{vis\_path}_i)
% \end{equation}
\noindent where $\text{vis}_i$ denote visual observation identifiers, $\text{vis\_tag}_i$ is a semantic label associated with the visual observation.

\textbf{Semantic Memory.} 
In addition to encounter-level interactions and observation-level perceptual evidence, clinical reasoning relies on an accumulated abstraction of a patient's prior diagnostic trajectory. To represent this longitudinal patient-specific state, MSM-Mem maintains a semantic memory module that aggregates historical interaction outcomes into a compact diagnostic summary. In our implementation, semantic memory maintains a per-patient longitudinal abstraction of prior diagnostic outcomes through deterministic aggregation of recent episodic impressions: $M_S^p(t) =
\mathcal{U}_S
\big(
M_S^p(t-1), a(t), d(t)
\big)$, 
% \begin{equation}
% M_S^p(t) =
% \mathcal{U}_S
% \big(
% M_S^p(t-1), a(t), d(t)
% \big),
% \end{equation}
where $\mathcal{U}_S(\cdot)$ denotes a patient-specific update operator that integrates the current impression $a(t)$ and decision outcome $d(t)$ into the existing semantic state $M_S^p(t-1)$, producing an updated representation of the patient’s evolving diagnostic profile.
\subsection{Memory-Conditioned Inference}
\label{sec:inference}
Given a new clinical observation $x^p(t)$ in step $t$ for patient $p$, MSM-Mem performs memory-conditioned inference by first retrieving relevant episodic and visual evidence, while incorporating patient history from semantic memory to construct a structured reasoning prompt. Specifically, we first compute a retrieval key, $z^p(t) = f_\theta(x^p(t)) / \|f_\theta(x^p(t))\|_2$
% \begin{equation}
%     z^p(t) = f_\theta(x^p(t)) / \|f_\theta(x^p(t))\|_2
% \end{equation}
\noindent Where $f_\theta(\cdot)$ is a visual encoder. Episodic and visual memory are queried via cosine similarity to identify relevant experience- and perception-level evidence:
$\mathcal{R}_k(z^p(t))
=
\operatorname{TopK}_{m_i \in M_k^p}
\big(
z^p(t)^\top \hat{z}_i
\big)$, where $k \in \{E,V\}$.

% \begin{equation}
% \mathcal{R}_k(z^p(t))
% =
% \operatorname{TopK}_{m_i \in M_k^p}
% \big(
% z^p(t)^\top \hat{z}_i
% \big),
% \quad
% k \in \{E,V\}.
% \end{equation}

Semantic memory, representing a patient-specific longitudinal abstraction, is incorporated as contextual state: $\mathcal{R}_S(p) = M_S^p$.
% \begin{equation}
% \mathcal{R}_S(p) = M_S^p
% \end{equation}
The retrieved memory components and current clinical query $Q(t))$ are subsequently fused into a structured prompt: $\mathcal{P}(t) =\text{Fuse}(M_S^p,\;
\mathcal{R}_E(z^p(t)),\;
\mathcal{R}_V(z^p(t)),\;
Q(t))
).$
% \begin{equation}
% \mathcal{P}(t) =\text{Fuse}(M_S^p,\;
% \mathcal{R}_E(z^p(t)),\;
% \mathcal{R}_V(z^p(t)),\;
% Q(t))
% ).
% \end{equation}

Inference is then performed using a MLLM backbone:

\begin{equation}
y_t =
\text{MLLM}
(\mathcal{P}(t),\; x^p(t)).
\end{equation}

Note that during inference, only the current observation is encoded by the visual backbone, while retrieved memory items are injected as textualized contextual evidence.

\subsection{Outcome-Driven Memory Update}
\label{sec:update}
Repeated clinical encounters may not always provide diagnostically novel information. Uncontrolled memory accumulation can therefore introduce redundant interaction traces and degrade retrieval quality over time. To maintain a compact and informative memory representation, MSM-Mem selectively performs memory updates based on multiple criteria. First, a newly observed interaction is retained only if its perceptual representation differs sufficiently from previously stored experiences:
$\max_{i}
\big(
z^p(t)^\top \hat{z}_i
\big)
<
\tau$,
% \begin{equation}
% \max_{i}
% \big(
% z^p(t)^\top \hat{z}_i
% \big)
% <
% \tau,
% \end{equation}
where $\tau$ denotes a predefined novelty threshold and $\hat{z}_i$ are normalized stored embeddings in episodic or visual memory. Second, the generated response must satisfy a minimum length requirement of $k$ words. Third, the generated impression is required to contain at least one clinically relevant keyword. If all criteria are satisfied, the episodic and visual memory stores for patient $p$ are updated. In contrast, the semantic memory store is updated after every interaction via Eq.~(8) to maintain an up-to-date abstraction of the patient’s evolving diagnostic state.

\section{Experiments \& Results}
\subsection{Dataset}
We evaluate MSM-Mem on a manually curated MIMIC-CXR \cite{johnson_mimic-cxr_2019} dataset comprising chest radiographs paired with corresponding radiology reports, clinical captions, and reasoning-oriented question–answer annotations. The dataset is partitioned at the subject level into disjoint subsets, with 30\% allocated for memory construction and updating to simulate continuous interaction accumulation, 60\% for fine-tuning baseline models for comparison, and 10\% for validation.
Within the memory split, 4\% is used to initialize the base memory, and the performance is probed at successive 6\%, 6\%, 6\%, and 8\% partitions to assess the impact of progressive memory updates.

\subsection{Implementation Details}

We adopt MoE-LLaVA~\cite{lin_moe-llava_2024} as the MLLM backbone. Baseline comparisons are conducted using the MoE-LLaVA base, fine-tuned (FT), and MoE-layered fine-tuned (MoE-FT) variants. The backbone used for visual embeddings extraction is the ImageBind~\cite{girdhar_imagebind_2023}. For memory retrieval, the Top-$k$ parameter is set to $k=10$ (in Appendix, we show for $k = 5$ as an example). The perceptual novelty threshold for memory write-back is set to $\tau = 0.9$. The experiments were conducted on 8$\times$ Nvidia A5000 GPUs.

\subsection{Effectiveness of Structured Memory}

Table~\ref{tab:tab1} presents the impression-level performance under different structured memory configurations. Across all model variants, incorporating any individual memory module yields consistent improvements over the memory-free baseline, indicating the complementary role of interaction- and perception-level contextualization in clinical reasoning. Among single-module configurations, episodic memory contributes the largest performance gain, suggesting that prior encounter-level decision traces provide highly informative experiential context for downstream impression generation. Visual memory also improves performance by enabling retrieval of perceptually similar imaging evidence. Notably, combining episodic and visual memory results in a substantial performance boost, highlighting the benefit of jointly leveraging decision trajectories and perceptual prototypes. The inclusion of semantic memory further enhances performance, with the full memory configuration (E+V+S) achieving the highest F1 and EM scores across all settings, demonstrating the effectiveness of integrating patient-level longitudinal state with prior experiential and perceptual evidence.

\begin{table}[t]
\centering
\caption{\textbf{Impression-level performance under different structured memory configurations.}
Average F1 (\%) and EM (\%) scores at the final writeback (WB4) phase.}
\label{tab:tab1}

\renewcommand{\arraystretch}{1.18}
\setlength{\tabcolsep}{3pt}

\begin{tabular*}{\textwidth}{@{\extracolsep{\fill}}ccc cc cc cc@{}}
\toprule
\multirow{2}{*}{\makecell{\textbf{Semantic}\\\textbf{Memory}}} &
\multirow{2}{*}{\makecell{\textbf{Episodic}\\\textbf{Memory}}} &
\multirow{2}{*}{\makecell{\textbf{Visual}\\\textbf{Memory}}} &
\multicolumn{2}{c}{\textbf{Base}} &
\multicolumn{2}{c}{\textbf{FT}} &
\multicolumn{2}{c}{\textbf{MoE-FT}} \\
\cmidrule(lr){4-5}\cmidrule(lr){6-7}\cmidrule(lr){8-9}
& & &
\textbf{F1$\uparrow$} & \textbf{EM$\uparrow$} &
\textbf{F1$\uparrow$} & \textbf{EM$\uparrow$} &
\textbf{F1$\uparrow$} & \textbf{EM$\uparrow$} \\
\midrule

$\times$ & $\times$ & $\times$ & 11.50 & 9.91 & 15.36 & 13.46 & 18.43 & 16.37 \\
$\checkmark$ & $\times$ & $\times$ & 13.33 & 11.29 & 18.70 & 16.68 & 25.69 & 23.77 \\
$\times$ & $\checkmark$ & $\times$ & \textbf{14.73} & \textbf{12.89} & \textbf{20.51} & \textbf{18.78} & \textbf{28.53} & \textbf{26.38} \\
$\times$ & $\times$ & $\checkmark$ & 14.16 & 12.24 & 19.59 & 17.85 & 27.25 & 25.44 \\
$\checkmark$ & $\checkmark$ & $\times$ & 18.35 & 16.19 & 30.76 & 28.39 & 36.42 & 34.68 \\
$\checkmark$ & $\times$ & $\checkmark$ & 18.41 & 16.21 & 29.69 & 27.53 & 36.59 & 34.71 \\

$\times$ & $\checkmark$ & $\checkmark$ &
\textbf{\textcolor{orange}{22.81}} & \textbf{\textcolor{orange}{20.37}} &
\textbf{\textcolor{orange}{37.64}} & \textbf{\textcolor{orange}{34.75}} &
\textbf{\textcolor{orange}{44.80}} & \textbf{\textcolor{orange}{42.39}} \\

$\checkmark$ & $\checkmark$ & $\checkmark$ &
\textbf{\textcolor{green!60!black}{27.62}} & \textbf{\textcolor{green!60!black}{25.43}} &
\textbf{\textcolor{green!60!black}{47.67}} & \textbf{\textcolor{green!60!black}{44.57}} &
\textbf{\textcolor{green!60!black}{54.92}} & \textbf{\textcolor{green!60!black}{51.50}} \\

\bottomrule
\end{tabular*}
\end{table}

We further evaluate the impact of structured memory across different MLLM backbones, including the MoE-LLaVA base, fine-tuned (FT), and MoE-layered fine-tuned (MoE-FT) variants. While stronger backbones consistently outperform their weaker counterparts in the absence of memory, all models exhibit substantial performance gains when augmented with structured memory modules. In particular, the MoE-FT backbone achieves the highest overall performance when combined with full memory (E+V+S), indicating that experience-informed contextualization remains beneficial even for more capable multimodal reasoning models. These results suggest that memory-augmented inference provides complementary gains beyond backbone capacity.

% \begin{figure}[t]
%   \includegraphics[width=1\linewidth]{Fig-6-final.png} \hfill
%   \caption {Impression-level F1 (\%) performance across successive memory write-back phases (Base, WB1–WB4) under different structured memory configurations, including episodic memory (E), episodic + visual memory (E+V), and episodic + visual + semantic memory (E+V+S), evaluated on MoE-LLaVA Base (a), Fine-tuned (b), and MoE Fine-tuned (c) backbones.}
%   \label{fig:fig2}
% \end{figure}

% \begin{figure}[ht]

% \includegraphics[width=\textwidth]{Fig-3.final.png},
% \caption{Impression-level F1 (\%) performance across successive memory write-back phases (Base, WB1–WB4) under different structured memory configurations, including episodic memory (E), episodic + visual memory (E+V), and episodic + visual + semantic memory (E+V+S), evaluated on MoE-LLaVA Base (a), Fine-tuned (b), and MoE Fine-tuned (c) backbones.} \label{fig1}
% \label{fig:fig2}
% \end{figure}

% \begin{figure}[t]
%   \includegraphics[width=1\linewidth]{Fig-4.png} \hfill
%   \caption {The impact of continuous memory improvement with writebacks (from base memory to WB4).}
%   \label{fig:fig3}
% \end{figure}

% \begin{figure}[h]
% \includegraphics[width=\textwidth]{Fig-8.1.png}
% \caption{The impact of continuous memory improvement with writebacks (from base memory to WB4).}  \label{fig1}
% \label{fig:fig3}
% \end{figure}

\subsection{Memory Scaling Analysis}

Figure \ref{fig:fig3} denotes the effectiveness of longitudinal clinical querying at best memory combinations (E, E+V, S+E+V) for the MoE fine-tuned checkpoint. After adding full memory (S+E+V), the agent facilitates its semantic, episodic, and visual experts to efficiently infer on the current query with the accumulation of prior queries.

Figure~\ref{fig:fig4} presents a representative example illustrating the effect of progressive memory write-back on longitudinal clinical reasoning. With static base memory (WB0), the model fails to contextualize new imaging findings. As episodic (WB1), multimodal structured (WB2), and region-level grounded (WB3) memory are incrementally incorporated, the model increasingly captures temporal progression and spatial correspondence between studies. The full memory configuration (WB4) further integrates retrieval history and prior query traces, enabling clinically grounded interpretation of progressive right lower lobe consolidation consistent with disease progression.

However, the MoE fine-tuned (MoE-FT) checkpoints over WB4 memory phase and full memory combination (S+V+E) perform the best among all experimental cases with a maximum (F1) accuracy of ~55\%, with 27.3\% and 26.07\% (gain) increase than its prior model (marked with green text for F1 impression average) (refer to Figure \ref{fig:fig5}).

Figure~\ref{fig:fig6} shows the impact of progressive memory accumulation across successive phases (Base, WB1–WB4). Performance consistently improves as more memory is incorporated, indicating that incremental interaction-derived evidence enhances downstream reasoning. While episodic memory alone yields modest gains, combining episodic and visual memory (E+V) leads to more substantial improvements. The full memory configuration (E+V+S) achieves the highest performance and exhibits the steepest growth across phases, suggesting that patient-level semantic context becomes increasingly beneficial as longitudinal interaction history accumulates.

However, the impact of non-MoE gating (regular FT) vs, MoE-FT at the final phase (WB4) for all memory combinations are displayed in Figure \ref{fig:fig7}. It showed that episodic (E), visual (V), episodic+ visual (E+V), and full (S+E+V) has got maximum gain of ~8\%, 7.66\%, 7.16\%, and 7.25\% under the MoE-FT gating network compare to its non-MoE version, which verifies the logic towards highest performers in Table \ref{tab:tab1}.

\section{Limitations and Potential Risks}
Even though the proposed agent memory surpasses state-of-the-art approaches, we still need to consider a few factors in real-life biomedical applications. First, the memory may cause hallucinations if it encounters irrelevant prior cases, outdated patient states, partially matched embeddings, and noisy episodic embeddings. In that scenario, incorrect reasoning, fabricated continuity, and clinically unsafe recommendations can occur. Secondly, during memory writeback, error propagation may occur. For instance, an incorrect writeback of the generated query might later amplify the error and cause long-term cascading hallucination. Next, clinical pattern changes are unpredictable. In some cases, current evidence contradicts older evidence in real life. In that scenario, expert human interventions are required heavily. Lastly, in some cases, misaligned latent spaces, unstable memory retrieval, and a saturated writeback window could occur.
\section{Code and Data Availability}
All experiments use a publicly available dataset: MIMIC-CXR \cite{johnson_mimic-cxr_2019}. Our code, preprocessing scripts, and data splits are available at
\url{https://github.com/asad14053/Bio-Med-Agent}.
\section{Conclusion}
We propose MSM-Mem, a medical structured multimodal memory framework that enables experience-informed clinical reasoning by decomposing interaction-derived evidence into episodic, visual, and semantic memory. By progressively accumulating longitudinal interactions and perceptual evidence, MSM-Mem allows MLLM to condition inference on prior decision trajectories and patient-specific diagnostic state. Experiments on a curated MIMIC-CXR benchmark demonstrate consistent performance gains across model backbones, with further improvements observed as memory is incrementally updated. Future work will explore multimodal extension to additional clinical data sources to further enhance the generalizability of memory-conditioned clinical reasoning.

\begin{ack}
The authors received no specific funding for this work. The authors declare no
competing interests.
\end{ack}

\bibliographystyle{plainnat}
\bibliography{references}

\appendix
\section{Supplementary Details}
\label{sec:appendix}

\subsection{MSM Architecture}

We propose MSM-Mem multimodal architecture designed for multimodal semantic insights, temporal cues, and long-term reasoning in biomedical applications. The pipeline consists of 5 stages including offline, online, efficient retrieval and adaptive routing loop, and query inference (refer to Figure 2). The offline loop is divided into multimodal encoders, memory index builder, multimodal base memory construction comprises of semantic, episodic, and visual features followed by sequential writebacks. The base memory in offline loop is built over a large heterogeneous dataset and built only once during system setup .The online loop comprises of unified retriever, prompt builder, reasoning model with adaptive routing. The online loop is gradually enhanced world memory time to time if new heterogeneous data occurred. This ensure that, the inference becomes almost identical to ground truth incrementally through continuous write back phases. However, the retrieval consists of relevant memory modality, searches previous episodic embedding, re-ranking, and efficient fusion module. And, the adaptive routing chooses mixture of expert (MoE) efficiently based on ranking and cosine-similarity of memory features. Finally, each query is fed into multimodal foundation agent by utilizing retrieved memory, cue history and prompts to generate final response or, report. An intensive ablation study is done to show the performance of the proposed agent 
under various combinations of models and memory embedding.

\subsubsection{Offline Loop}
The offline loop is only executed once while forming the base memory for the first time. Steps of offline loop can be described as-

\textbf{(a) Preprocessing}: The base memory creation starts with data preprocessing into memory items. We assume a deterministic operator $\Phi_{\text{pre}}$ for preprocessing that converts a multimodal clinical dataset 
$D = \{d_i\}_{i=1}^{N}$, where $d_i$ denotes individual data entries, into a set of memory items $M$:

\begin{equation}
M = \{m_i\}_{i=1}^{N} = \Phi_{\text{pre}}(D)
\label{eq:preprocessing}
\end{equation}

Here, schema initialization and consistent IDs are assigned to each $d_i$, while invalid or unsupported entries are discarded. Thus, the final memory item is represented as:

\begin{equation}
\begin{aligned}
m_i = \Big(
&\text{item\_id}_i,\;
\text{subject\_id}_i,\;
\text{media\_type}_i, \\
&f_i^{\text{key}},\;
F_i,\;
\text{metadata}_i
\Big)
\end{aligned}
\label{eq:memory_item}
\end{equation}

If the input is a video, then $f_i^{\text{key}}$ denotes the deterministically selected keyframe extracted from the video; otherwise, it corresponds to the image itself. The term $F_i$ denotes the associated file path.

\textbf{(b) Base Memory builder and Creation}: The memory builder module creates three memory representations:
episodic, visual, and semantic. Let us assume a query with
identifier $p$ (subject/patient ID) and an associated reference
frame $x_i$. For filtered memory items,
$M = \{m_i\}_{i=1}^{N}$, the framework employs the
ImageBind vision encoder~\cite{girdhar_imagebind_2023} with $224 \times 224$
resolution, CLIP-style mean/std normalization, and L2
normalization to construct a unified latent space:

\begin{equation}
z_q = \mathrm{norm}\!\left(f_{\theta}(m_i)\right)
\label{eq:latent_space}
\end{equation}

where
$z_q \in \mathbb{R}^{d}$,
$f_{\theta}(\cdot)$ denotes the encoder function,
and the embedding dimension is fixed at $d = 1024$.

The builder then applies cosine similarity for future retrieval
prior to storage indexing. As a result, the storage key is
represented as a unit vector $\hat{z}_i$ following the
FAISS~\cite{douze_faiss_2025} indexing scheme. After L2 normalization,
FAISS indexing becomes equivalent to cosine similarity:

\begin{equation}
\hat{z}_i =
\frac{Z_i}
{\left\| Z_i \right\| + \epsilon}
\label{eq:faiss_norm}
\end{equation}

where $\epsilon$ denotes the normalization stability constant.

\textbf{(c) Episodic Memory Construction}:
Each query associated with a timestamp contributes to an
episodic index using the same key $\hat{z}_i$, linked with
episodic memory metadata:

\begin{equation}
\small
\begin{aligned}
\mathrm{metadata} = \{
&\text{episode\_id},\;
\text{patient\_id},\;
\text{timestamp},\;
z, \\
&\text{question},\;
\text{answer\_short},\;
\text{decision}, \\
&\text{media\_path},\;
\text{clip\_desc}
\}
\end{aligned}
\label{eq:episodic_metadata}
\end{equation}

The memory system employs a uniform scalar-product catalog
(\texttt{IndexFlatIP}) and stores coupled aligned attributes
within the same directory structure:
\texttt{episodic.jsonl} (metadata) and
\texttt{episodic.faiss} (latent vectors).

Notably, episodic alignment is maintained through timestamp
ordering and insertion sequence. The $i$-th latent vector added
to the FAISS index directly corresponds to the $i$-th JSONL
entry stored in \texttt{episodic.jsonl}, enabling a traceable
mapping from retrieved indices to human-interpretable memory
resources.

\textbf{(d) Visual Memory Construction}: Simultaneously, the same matched key $\hat{z}_i$ is used as
input to a FAISS-based visual index (\texttt{IndexFlatIP}),
which maps to a specific reference-frame file path, the
originating episodic item (\texttt{source\_episode\_id}),
and optional labels such as imaging view
(\texttt{visual\_tag}).

The visual memory is stored as
\texttt{visual.jsonl} (metadata) and
\texttt{visual.faiss} (visual embeddings), which are aligned
through insertion order.

The visual memory metadata is defined as:

\begin{equation}
\small
\begin{aligned}
\mathrm{metadata} = \{
&\text{visual\_id},\;
\text{patient\_id},\;
\text{timestamp},\;
z, \\
&\text{media\_path},\;
\text{visual\_tag}, \\
&\text{source\_episode\_id}
\}
\end{aligned}
\label{eq:visual_metadata}
\end{equation}

\textbf{(e) Semantic Memory Construction}: 
Semantic memory is maintained individually for each category
under per-patient JSON files, such as
\texttt{<category\_id>} within
\texttt{semantic/<patient\_id>.json}.

For every biomedical cue, the memory builder retrieves the
\textit{Impression} section from the ground-truth training
report and appends it to a bounded history buffer
($k_{\mathrm{sem}} = N$, where the default value is $N=5$)
to generate a deterministic patient-level semantic summary
sequence used for contextual prompting:

\begin{equation}
\mathrm{Semantic}(\mathrm{patient})
=
\mathrm{update}
\big(
\mathrm{Impression}(\mathrm{report})
\big)
\label{eq:semantic_update}
\end{equation}

This semantic memory is strictly restricted to
case-specific semantic contexts. Therefore, it does not
perform decision-making or statistical aggregation; instead,
it preserves only temporally aligned impression summaries
and associated time markers.

\textbf{(f) Per-memory retrieval, Re-ranking, and Fusion}: The memory retrieval module is directly governed by ablation
tags, including semantic (\texttt{use\_sem}),
episodic (\texttt{use\_epi}), and visual
(\texttt{use\_vis}) memory selection through a learned router.

If semantic memory is activated, retrieval is performed using:

\begin{equation}
\mathrm{semantic.get}(p)
\label{eq:semantic_get}
\end{equation}

where $p$ denotes the subject/patient identifier. This
function returns a patient-specific JSON state containing a
bounded and continuously updated impression history stored in
\texttt{patient\_summary} (default: three fields).

If episodic memory retrieval is enabled, the framework uses:

\begin{equation}
\mathrm{episodic.query}
\left(
z_q,\;
k_{\mathrm{epi}},\;
p
\right)
\label{eq:episodic_query}
\end{equation}

where the default value is $k_{\mathrm{epi}} = 5$.

Similarly, visual memory retrieval is defined as:

\begin{equation}
\mathrm{visual.query}
\left(
z_q,\;
k_{\mathrm{vis}},\;
p
\right)
\label{eq:visual_query}
\end{equation}

where the default value is $k_{\mathrm{vis}} = 5$.

Both memory modules internally sample up to $5$K items and
filter them using $p = \text{patient\_id}$ to preserve
patient-consistent retrieval constraints. Consequently, the
final top-$k$ memory candidates are returned in descending
FAISS similarity ranking order.

Let the retrieved memory item be denoted as
$M_i$, where:

\[
M_i \in \{M_E,\; M_S,\; M_V\}
\]

corresponding to episodic, semantic, and visual memories,
respectively. Each retrieved item generates an associated
query cue $q_i$ together with the batch of previously
retrieved histories:

\[
r_{N-1} =
\{r_1,\; r_2,\; \dots,\; r_{N-1}\}
\]

for the prior $(N-1)$ retrieval steps. The retrieval agent
then outputs either a memory-query pair or a STOP command.
Therefore, for a query $q$, the retrieval policy is defined
as:

\begin{equation}
R(q, r_{N-1}) =
\begin{cases}
(M_i, q_i), & i < N \\
\text{STOP}, & \text{otherwise}
\end{cases}
\label{eq:retrieval_policy}
\end{equation}

\textbf{(g) Re-ranking}: 
The retrieved memory, after incorporating newly merged data,
is determined purely through cosine-similarity ranking using
FAISS (\texttt{IndexFlatIP}) with L2-normalized embeddings.
Consequently, higher similarity scores are ranked first.

The re-ranking score is computed as:

\begin{equation}
\mathrm{ranking\_score}
=
\hat{z}_q^{\,T}\hat{z}_i
\label{eq:ranking_score}
\end{equation}

where $\hat{z}_q$ and $\hat{z}_i$ denote the normalized query
and indexed latent embeddings, respectively.

Finally, the re-ranked memory items are filtered using
\texttt{patient\_id} consistency constraints and clipped to
the top-$k$ entries (default: $k < 10$ for memory-safe
retrieval) according to descending
\texttt{ranking\_score} values.

\textbf{(h) Memory Fusion}: 
The fetched memory items are transformed into an ordered and
bounded textual representation and assembled into the final
memory prompt:

\begin{equation}
\small
\begin{aligned}
\mathrm{build\_prompt}\{
&\mathrm{semantic},\;
\mathrm{episodic\_ping}, \\
&\mathrm{visual\_ping},\;
\mathrm{question},\;
\mathrm{clip\_desc}
\}
\end{aligned}
\label{eq:build_prompt}
\end{equation}

The fusion strategy follows a mostly static and chronological
ordering scheme:

\begin{enumerate}
    \item Instruction captions, section headers, and optional
    contextual remarks (\texttt{context\_note});
    
    \item Patient-specific semantic context, where
    \texttt{semantic.patient\_summary} is truncated to fewer
    than $800$ characters;
    
    \item Relevant longitudinal records retrieved from
    episodic memory;
    
    \item Related historical visual embeddings retrieved from
    visual memory.
\end{enumerate}

Within each retrieved block, memory entries are rendered
through the \texttt{\_to\_text} function. Episodic memory
entries are converted into \texttt{(Q:A)} pairs using
\texttt{question} and \texttt{answer\_short}, whereas visual
memory entries are transformed into
\texttt{<visual\_tag>} descriptors.

To maintain lightweight and budget-aware prompt fusion, each
memory block is restricted to at most two retrieved
components (\texttt{max\_items} $= 2$), and every component is
further truncated to a maximum of $800$ characters.

The final prompt concludes with a \texttt{Question} section
followed by a constrained response template requiring both
\texttt{Findings:} and \texttt{Impression:} outputs.

Notably, modality information is expressed indirectly through
section headers rather than explicit memory tags such as
\texttt{<semantic>}, \texttt{<episodic>}, or
\texttt{<visual>}. Similarly, retrieved histories are exposed
through \texttt{context\_note} abstractions instead of direct
retrieval logs.

Ultimately, the fusion stage depends on top-$k$ retrieval
selection (default:
$k_{\mathrm{epi}} = k_{\mathrm{vis}} = 5$) together with
strict upper bounds (two items per memory block) to minimize
duplication while preserving lineage through retained markers
and retrieval paths.

\textbf{(i) Metrics}:
The evaluation framework employs two primary metrics:
token-level F1 score and Exact Match (EM). These metrics are
computed over three evaluation settings:

\begin{enumerate}
    \item Full clinical report,
    \item \textit{Impression} section,
    \item \textit{Findings} section.
\end{enumerate}

The evaluation pipeline extracts the corresponding report
segments using the \texttt{extract\_section()} function and
then applies Exact Match (EM) and token-level F1 scoring to
the decoded outputs.

Formally, the evaluation metrics are represented as:

\begin{equation}
\mathrm{Metrics}
=
\{
\mathrm{F1}_{\mathrm{token}},
\;
\mathrm{EM}
\}
\label{eq:evaluation_metrics}
\end{equation}

The computed scores are accumulated across all evaluation
iterations to produce averaged performance statistics, such
as:

\[
\{
\mathrm{F1}_{\text{imp}},
\;
\mathrm{EM}_{\text{imp}}
\}
\]

Similar averaging is performed for both the
\textit{Findings} and full-report evaluation settings during
the final evaluation run.

\textbf{(j) Writeback}: Memory writeback is executed for each generated item and is
regulated by three gating conditions.

First, novelty is computed only against retrieved episodic
memory entries for the incoming query $q$ using the highest
(top-1) FAISS cosine-similarity score
$S_{\mathrm{highest}}$:

\begin{equation}
\mathrm{novelty}(q)
=
\mathbb{1}
\left[
S_{\mathrm{highest}} < \tau
\right]
\label{eq:novelty}
\end{equation}

where $\tau$ denotes the novelty threshold
(default: $\tau = 0.90$).

Second, the generated \textit{Impression} segment must exceed
a minimum length constraint:

\begin{equation}
|\mathrm{Impression}|
\geq k
\label{eq:min_length}
\end{equation}

where the default value is $k = 120$ words.

Third, keyword-based gating ensures clinical relevance and
context quality. The generated \textit{Impression} text must
contain at least one keyword from both the
\texttt{writeback\_change\_keywords} list
(e.g., ``severe'', ``worse'', ``new'') and the
\texttt{writeback\_finding\_keywords} list
(e.g., ``effusion'', ``reflux''). These checks are applied
through the \texttt{has\_any\_keyword()} function.

Target samples selected for writeback are regulated by
memory-specific and per-patient constraints. Writeback is
blocked whenever a patient reaches the episodic or visual
memory capacity limit
(\texttt{writeback\_k\_epi},
\texttt{writeback\_k\_vis}),
which defaults to $200$ items per subject for memory safety.

Visual-memory writeback can additionally be disabled using
the \texttt{writeback\_no\_visual} flag. Furthermore,
writeback operations may be restricted globally per run
(\texttt{writeback\_max}) or per-patient/per-run
(\texttt{writeback\_max\_per\_patient}).

Once all writeback conditions are satisfied, the framework
executes:

\begin{enumerate}
    \item Semantic update:
    \texttt{semantic.update()} to the per-patient JSON file;
    
    \item Episodic update:
    \texttt{episodic.add()} to the JSONL and FAISS index;
    
    \item Visual update:
    \texttt{visual.add()} to the JSONL and FAISS index,
    associating the reference frame with a newly generated
    episode identifier.
\end{enumerate}

\subsubsection{Online Loop}

Online loop is normally executed during each training, testing, and inference cycle. The repeating steps in the online loop consist of a memory retriever, prompt builder, and evaluation (refer to Figure 2.), which uses the same wrapper classes from offline loops. The other steps in online loops can be explained as below-

\textbf{(a) Model Routing and Controlled Inference:}
The model router selects the inference wrapper class based on
checkpoint configuration settings. The function
\texttt{ckpt\_is\_moe()} inspects the
\texttt{config.json} file and redirects execution to
\texttt{MoELLaVAInfer} when a Mixture-of-Experts (MoE)
configuration is detected; otherwise, inference is performed
through \texttt{LlaVAPhiInfer}.

This mechanism separates standard and MoE-based execution
paths while preserving a unified retrieval and prompt-building
pipeline for both settings.

During inference, a strict context-budget constraint is
enforced:

\begin{equation}
\small
\mathrm{token\_budget}
=
\mathrm{context\_len}
-
\mathrm{max\_new\_tokens}
-
\mathrm{margin}
\label{eq:token_budget}
\end{equation}

Only the retrieved contextual segment of the input prompt is
compressed using the
\texttt{\_shrink\_user\_prompt()} function. This intermediate
clipping strategy preserves the instruction header, question
tail, and tail-priority fallback regions before applying
additional token-level truncation safeguards if required.

The decoding process is deterministic and bounded. Generation
terminates automatically through the
\texttt{KeywordsStoppingCriteria} mechanism based on dialogue
delimiter conditions, ensuring stable and fixed-format
clinical report generation.

\textbf{(b) MoE Router Gating and Adaptation:}
Mixture-of-Experts (MoE) adaptation is achieved by updating
only the router gating weights $W_g$ during fine-tuning while
keeping all expert parameters fixed.

Within the MoE block~\cite{linMoELLaVAMixtureExperts2024}, the router gating network
maps each hidden token representation to a mixture of expert
selection scores using the gating weights $W_g$. The routing
mechanism then dispatches computation to a compact subset of
experts (e.g., top-1 or top-2 experts depending on the MoE
configuration protocol).

Formally, the routing process can be represented as:

\begin{equation}
g(x) = \mathrm{TopK}(W_g x)
\label{eq:moe_router}
\end{equation}

where $x$ denotes the hidden token representation and
$g(x)$ represents the selected expert-routing scores.

By optimizing only the gating weights $W_g$, the reasoning
model learns an adaptive routing policy that dynamically
allocates computational resources to the most relevant
experts according to the clinical context.

The expert routing operation is executed internally through
the fine-tuned routing module during
\texttt{self.model.generate()} inference execution.

\subsection{Mixture of Expert (MoE) Gating}

If the framework contains $N$ experts,
the expert set is defined as:

\begin{equation}
E = \{E_1, E_2, \dots, E_N\}
\label{eq:expert_set}
\end{equation}

The gating controller selects experts from $E$
based on the routing score:

\begin{equation}
g =
\mathrm{Softmax}(W_g x)
\label{eq:gating_score}
\end{equation}

where $W_g$ denotes the gating-weight matrix learned during
MoE pre-training and $x$ represents the input embedding.

\section{Additional Figures}
\FloatBarrier
\renewcommand{\thefigure}{A\arabic{figure}}
\setcounter{figure}{0}

% \begin{figure}[t]
%   \includegraphics[width=1\linewidth]{Fig-A1.1}
%   \caption {Data preparation (JSON metadata) for medication adherence prediction (AdCare
% VLM).}
% \end{figure}

\begin{figure}[t]
  \centering
  \includegraphics[width=1\linewidth]{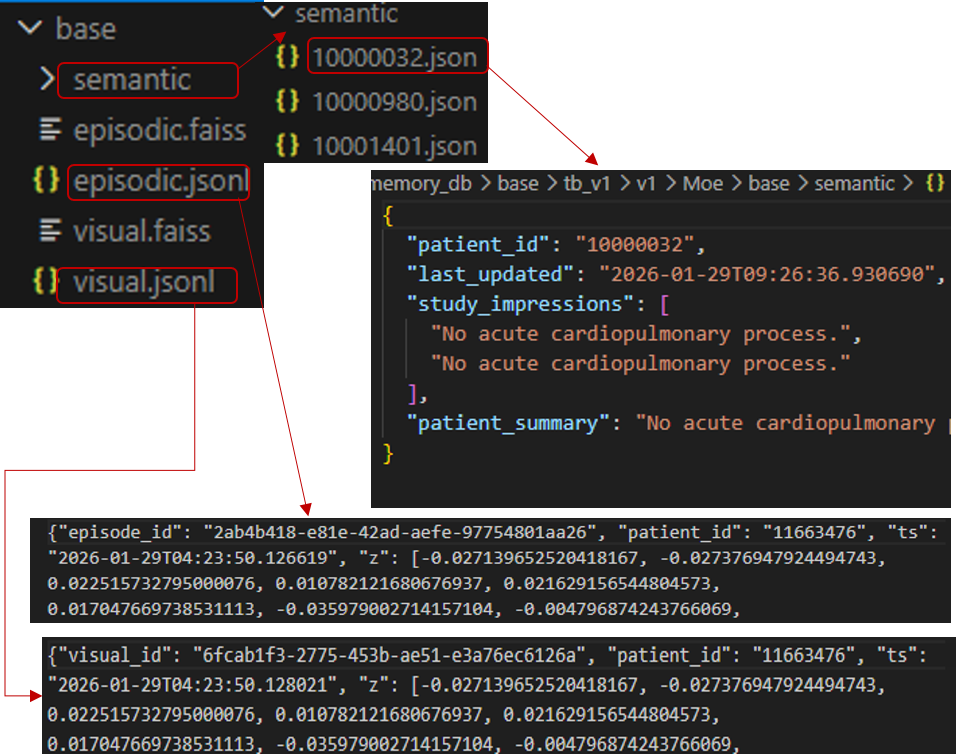}
  \caption {The real output of creating base memory comprises of semantic, episodic, and visual 
memory for each subject for our MSM-Mem model.}
\label{fig:figa1}
\end{figure}

\begin{figure}[t]
  \centering
  \includegraphics[width=1\linewidth]{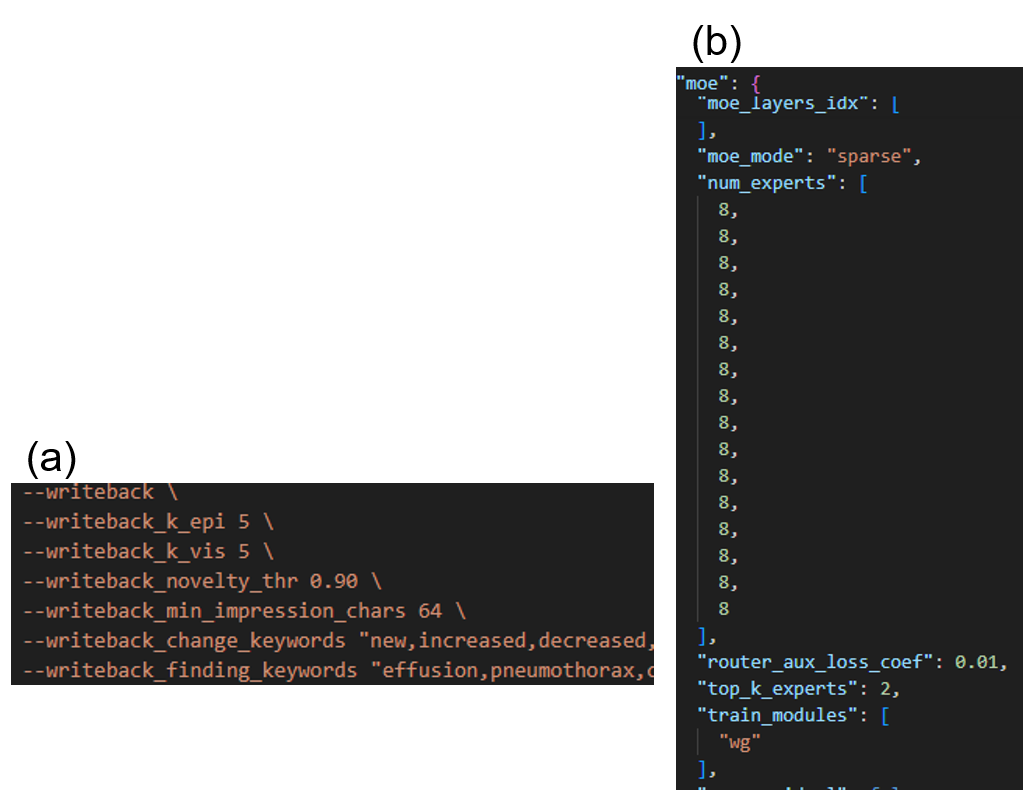}
  \caption { (a) The writeback controlling parameter, (b) MoE router gating condition for our MoE based MSM-Mem Agent model (Here, we utilize top k = 2 out of 8 due to system limitations during 
MoE routing).}
\label{fig:figa2}
\end{figure}

\end{document}